\documentclass[conference]{IEEEtran}
\usepackage{times}
\usepackage[T1]{fontenc}

\usepackage[numbers]{natbib}
\usepackage{multicol}
\usepackage{graphicx}
\graphicspath{ {./images/} }
\usepackage{subfigure}
\usepackage{float}
\usepackage{amsmath,amssymb}
\usepackage{amsfonts}
\usepackage{xcolor}
\usepackage[bookmarks=true]{hyperref}
\usepackage{cleveref}
\DeclareMathOperator*{\argmax}{argmax}
\usepackage[
top = 0.75in,
bottom = 0.8in,
left   = 0.75in,
right  = 0.75in]{geometry}

\begin{document}
\title{\vspace*{0.25in}
Robotic Table Tennis with Model-Free Reinforcement Learning}

\author{\authorblockN{Wenbo Gao\authorrefmark{1}\authorrefmark{4},
Laura Graesser\authorrefmark{1}\authorrefmark{2},
Krzysztof Choromanski\authorrefmark{1}\authorrefmark{2}, 
Xingyou Song\authorrefmark{2},
Nevena Lazic\authorrefmark{3},\\
Pannag Sanketi\authorrefmark{2},
Vikas Sindhwani\authorrefmark{2} and
Navdeep Jaitly\authorrefmark{5}}
\authorblockA{\authorrefmark{1}Equal contribution \authorrefmark{2}Robotics at Google \authorrefmark{3}DeepMind}
\authorblockA{\authorrefmark{4}Columbia University. Work done during Google internship \authorrefmark{5}Work done at Google}
}

\maketitle

\begin{abstract}
We propose a model-free algorithm for learning efficient policies capable of returning table tennis balls by controlling robot joints at a rate of 100Hz. We demonstrate that evolutionary search (ES) methods acting on CNN-based policy architectures for non-visual inputs and convolving across time learn compact controllers leading to smooth motions. 
Furthermore, we show that with appropriately tuned curriculum learning on the task and rewards, policies are capable of developing multi-modal styles, specifically forehand and backhand stroke, whilst achieving 80\% return rate on a wide range of ball throws. We observe that multi-modality does not require any architectural priors, such as multi-head architectures or hierarchical policies.
\end{abstract}


\section{Introduction}

Recent advances in machine learning (ML), and in particular \emph{reinforcement learning} (RL), have led to progress in robotics \cite{QTOPT2018CORL,tobin, james}, which has traditionally focused on optimal control. Two key advantages of ML are its ability to \emph{leverage increasing data and computation} and \emph{learn task-specific representations}. ML algorithms reduce the need for human knowledge by automatically learning useful representations of the data. For difficult problems, this is crucial because the complexity often exceeds what can be accomplished by explicit engineering. The effectiveness of ML is dependent on having large amounts of data. While this limits ML when little data is available, it becomes a strength because ML \emph{continues} to scale with ever-increasing amounts of data and computation: a system can be improved, often to the new state of the art, simply by gathering more data. These advantages make RL appealing for robotics, where large-scale data can be generated by simulation, and systems and tasks are often too complex to explicitly program.

In this paper, we apply RL to robotic table tennis. In contrast to some of the previous manipulation tasks solved by RL \cite{QTOPT2018CORL} where inference time of few hundred milliseconds is acceptable, table tennis is distinguished by the need for \emph{high-speed} perception and control, and the high degree of precision required to succeed at the task --- the ball needs to be hit very precisely in time and space. Given these challenges, prior work on robotic table tennis is typically model-based, and combines kinematics for predicting the full ball trajectory with inverse kinematics for planning the robot trajectory. Most recent research first identifies a \emph{virtual hitting point} \cite{Ramanantsoa94} from a partial ball trajectory and predicts the ball velocity, and potentially spin, at this point. A target paddle orientation and velocity is then determined, and a trajectory is generated to bring the paddle to the desired target at a particular time \cite{Miyazaki2006LearningTD, Muelling2009ACM, Muelling2012LearningSelectGen, Muelling2010LearningTTMOMP, Matsushima2005ALA, Huang2015LearningOS, Mahjourian2018HierarchicalPD}. Most systems include a predictive model of the ball (learned or analytical), and may also utilize a model of the robot. It is an open question whether end-to-end RL is an effective approach for robotic table tennis (and complex high-speed tasks in general), which motivates our approach.

We are also motivated by human play. When humans play table tennis, they exhibit a variety of stroke styles (multi-modality). These styles include forehand, backhand, topspin, backspin, etc. We are interested in understanding if multi-modal style emerges naturally during training, and if not, what techniques are required to generate this behavior.

\begin{figure}[t]
\centering
\includegraphics[width=0.45\textwidth]{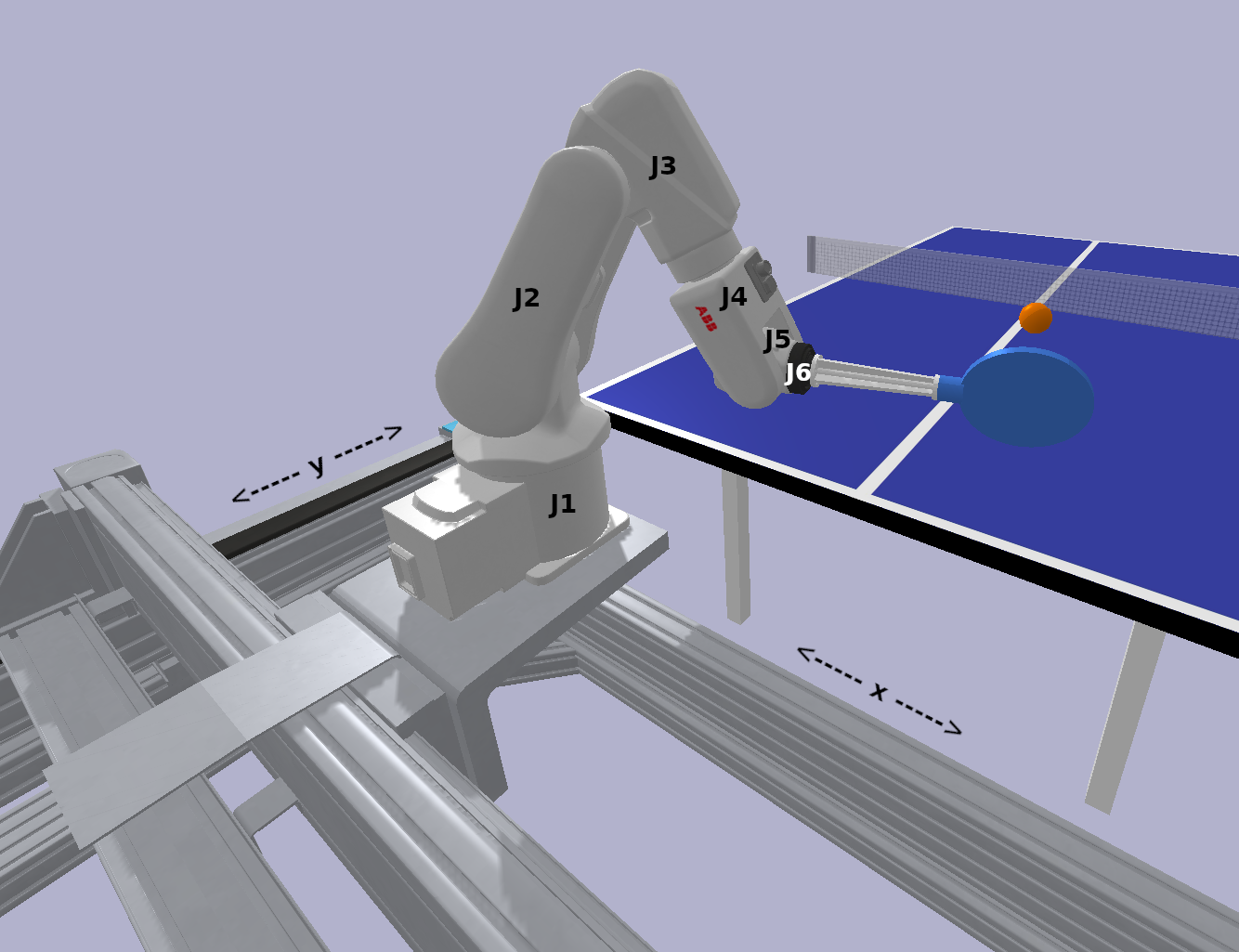}
\caption{Simulated table tennis 8 DOF robot: 6 DOF arm with revolute joints (labeled $\mathrm{j1}$-$\mathrm{j6}$) + 2 linear axes.}
\label{robot_arm}
\vspace{-6mm}
\end{figure}

In this work, we describe how to learn a multi-modal model-free policy to directly control a simulated table tennis robot in joint space using RL, without relying on human demonstrations, and without having a separate system to predict the ball trajectory. To simplify the task, we focus on forehand and backhand play styles without spin. Our policies take as input short histories of joint positions and ball locations, and output velocities per joint at 100Hz.

A video demonstrating our system can be viewed at \url{www.youtube.com/watch?v=-eHeq1nvHAE}. The video is divided into segments, each exhibiting a different policy (\emph{Policy A -- E}). Policy A is an example of a strong policy which returns \textbf{80\%} of randomly thrown balls to the opponent side of the table (see \Cref{tab:top_bimodal}), and exhibits high-level decisions through \emph{bimodal} (forehand and backhand) play. Policies B -- E in the video show ablation studies (\Cref{sec:architecture-results} to \ref{sec:action-filter}) and failure modes (\Cref{sec:landing_bonus}). Policies B and C illustrate the differences in style between CNN and MLP architectures.

In summary, our contributions are the following:
\vspace{-3mm}
\newline
\begin{itemize}
\item To the best of our knowledge, we train the first table tennis controller using model-free techniques without a predictive model of the ball or human demonstration data (\Cref{sec:model-free-policies}). 
\item We show that it is possible for a policy to learn multi-modal play styles with careful curriculum learning, but with no need of architectural priors (\Cref{sec:playstyles}).
\item We demonstrate that convolutions across time on non-visual inputs lead to smoother policies. They are also more compact than MLPs, which may be particularly beneficial for ES-methods (\Cref{sec:model-free-policies,sec:architecture-results}).

\end{itemize}
\section{Related Work}
Research in robotic table tennis dates back to 1983 when Billingsley \cite{Billingsley83} proposed a robotic table tennis challenge with simplified rules and a smaller table. Subsequently, several early systems were developed \cite{Knight1986PingpongplayingRC,Hartley87,Hashimoto1987DevelopmentOP}; see \cite{Muelling2010Biomem} for a summary of these approaches.
At the time of the last competition in 1993, the problem remained far from solved.

Standard model-based approaches to robotic table tennis, as discussed in the introduction, consist of several steps: identifying virtual hitting points from trajectories, predicting ball velocities,
calculating target paddle orientations and velocities, and finally generating robot trajectories leading to desired paddle targets.
\cite{Miyazaki2002RealizationOT, Miyazaki2006LearningTD, Anderson1988ARP, Muelling2010SimulatingHT, Zhu2018TowardsHL} take this approach and impose the additional constraint of fixing the intercept plane in the $y$-axis (perpendicular distance to the net). \cite{Huang2015LearningOS, Sun2011BalanceMG, Mahjourian2018HierarchicalPD} allow for a variable ball intercept, but still use a virtual hitting point. The predictive ball model is either learned from data \cite{Miyazaki2002RealizationOT, Matsushima2003LearningTT, Matsushima2005ALA, Miyazaki2006LearningTD} or is a parameterized dynamics model, which can vary in complexity from Newtonian dynamics with air drag \cite{Muelling2010Biomem, Muelling2010SimulatingHT}, to incorporating restitution and spin \cite{Zhu2018TowardsHL}. Robots vary from low-DOF robots with simple motion generation \cite{Miyazaki2002RealizationOT, Matsushima2003LearningTT, Matsushima2005ALA, Miyazaki2006LearningTD} to anthropomorphic arms with strong velocity and acceleration constraints \cite{Muelling2010Biomem, Muelling2010LearningTTMOMP, Tebbe2018ATT, Gao2019MarkerlessRP}.

Once a paddle target has been generated, the trajectory generation problem is still far from straightforward, especially if the robotic system has strong constraints. \cite{Muelling2010Biomem} resolve the redundancy in a 7DOF system by minimizing the distance to a `comfort posture' in joint space whilst finding a paddle position and orientation that coincides with the hitting point. \cite{Muelling2010LearningTTMOMP, Muelling2012LearningSelectGen} create a movement library of dynamical system motor primitives (DMPs) \cite{Ijspeert2002MovementIW} from demonstrations, and learn to select from amongst them and generalize between them with their Mixture of Motor Primitives (MoMP) algorithm.

\cite{Huang2016JointlyLT} and \cite{Ko2018OnlineOT} take a different approach and do not identify a virtual hitting point. \cite{Huang2016JointlyLT} use a combination of supervised and reinforcement learning to generate robot joint trajectories in response to ball trajectories. An important component of this system is a map which predicts the entire ball trajectory given the initial ball state estimated from a collection of measured ball positions. \cite{Ko2018OnlineOT} use three ball models (flight model, ball-table rebound model, ball-racket contact) to generate a discrete set of ball positions and velocities, given ball observations queried from the vision system. Desired racket parameters for each set of ball positions are generated and then the optimization for robot joint movements is run. The system is fast enough to generate trajectories online.

\cite{Huang2016JointlyLT} and \cite{Ko2018OnlineOT} are the closest to our work. We use a similar anthropomorphic robotic system, a 6 DOF arm with revolute joints + 2 DOF linear axes, and do not make use of a virtual hitting point. However in our approach we do not use a predictive model of ball, nor do we use demonstrations to learn to generate trajectories.

To the best of our knowledge, two classes of model-free approaches have been applied to robotic table tennis. \cite{Zhu2018TowardsHL} frames the problem as a single-step bandit and uses DDPG \cite{LHPHETSW2016ICLR} to predict the linear velocity and normal vector of the paddle given the position, linear and angular velocity of the ball at the hitting plane in simulation. \cite{Akrour2016ModelFreeTP} learns a local quadratic time-dependent $Q$-function from trajectory data, which they use to optimize a time-dependent stochastic linear feedback controller. The initial policy is learned from demonstration data. By contrast the main focus of this paper is on-policy methods, and our policies produce temporally extended actions at 100Hz in joint space.
\section{Preliminaries}
\subsection{Robotic Table Tennis}\label{sec:rtt_background}
Our goal in this paper is to train policies to solve the basic task of returning balls launched from the opponent side of the table. Beyond this, we are also interested in the \emph{style} of the robot's play.

\textit{How} a policy acts is as important as \textit{how many} balls it can return, especially if the policy is used to control a physical robot. Good style -- smooth control operating within the robot's limits -- is crucial, in addition to the success rate.

Our policies should be able to execute complex playing styles involving high-level decision making, as humans do. In the context of robotic table tennis, an instance of this is \emph{bimodal play}: the ability of the policy to select between \emph{forehand} and \emph{backhand} swings. Moreover, this should be \emph{extensible}; in addition to the aforementioned goals of smoothness and bimodal play, the policy should have avenues for incorporating further styles or strategies as the difficulty of the task increases.

A key contribution of our paper is the development of methods to accomplish these goals using RL. We describe these methods in \Cref{sec:methods}.

\subsection{RL Background}
To describe RL policies and algorithms, we use the formalism of the \emph{Markov Decision Process} (MDP), which consists of a state space $\mathcal{S}$, action space $\mathcal{A}$, a reward function $\mathcal{R}:\mathcal{S}\times\mathcal{A}\to\mathbb{R}$, and stochastic transition dynamics $p(s_{t+1}|s_{t}, a_{t})$. We parameterize a policy $\pi:\mathcal{S}\to\mathcal{A}$ as a neural network with parameters $\theta\in\Theta$, denoted as $\pi_\theta$. The goal is to maximize the expected total reward $F(\theta) := \mathbb{E}_{a_t \sim \pi_\theta(s_t)} \left[\sum_{t=1}^{H} r(s_{t}, a_{t})\right]$, where $a_t \sim \pi_\theta(s_t)$ indicates that actions follow $\pi_\theta$.

Our controller $\pi_{\theta}$ generates continuous velocity commands in joint space. Within model-free RL, there are two broad classes of algorithms: those based on \emph{value functions}, with the canonical example being $Q$-learning, and those using \emph{direct policy search} \cite{SB1998MIT}. $Q$-learning, which learns a function $Q(s,a)$ to predict the expected reward starting at state-action pair $(s,a)$, has been successfully applied to manipulation problems \cite{QTOPT2018CORL}. Its disadvantage is that inference (i.e. selecting an action) involves solving an optimization problem (i.e. $\pi_{\theta}(s) = \argmax_{a} Q(s, a)$) which can take several hundred milliseconds for continuous action spaces, as in \cite{QTOPT2018CORL} (using $\mathrm{CMA}$-$\mathrm{ES}$). This is impractical for high-speed tasks like table tennis. Instead, we opt for direct policy search methods, which learn a mapping from states to actions.
\section{Methods: Applying RL to Table Tennis}\label{sec:methods}

In this section, we describe our key design choices for successfully applying RL to robotic table tennis: the policy architecture, the use of curriculum learning, and the choice of optimization algorithm.

\subsection{Policy Representation}\label{sec:general_policy_arch}
\begin{figure}[t]
\centering
\includegraphics[width=0.50\textwidth]{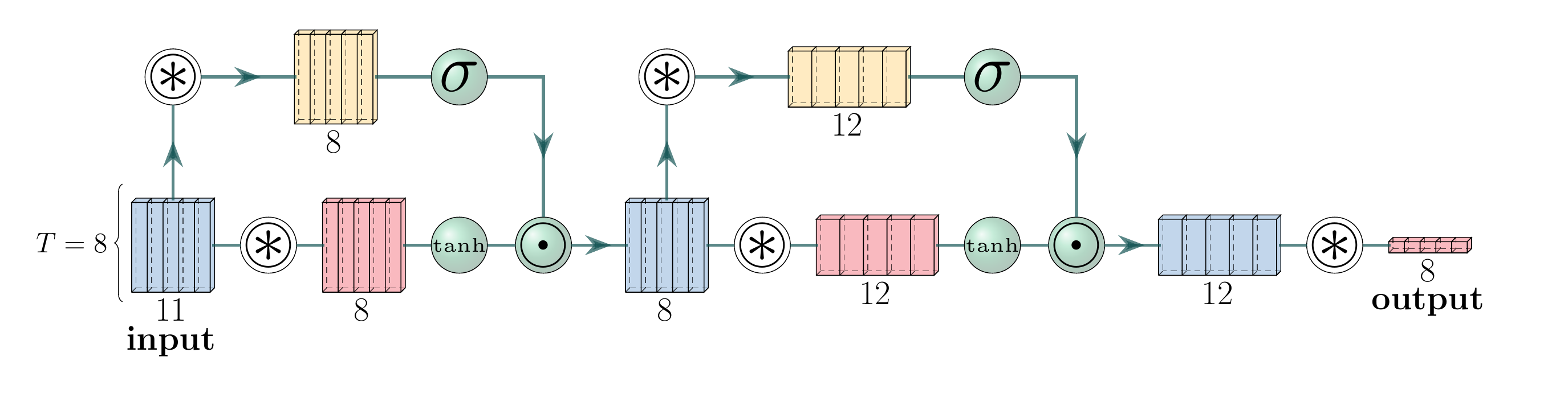}
\caption{Gated CNN architecture\protect\footnotemark. Rectangles represent channels, and 1D filters are applied on the vertical (time) dimension. Dilation reduces the height in successive layers. The symbol $\circledast$ denotes the application of convolutions, $\sigma$ denotes elementwise sigmoid, and $\odot$ denotes elementwise (Hadamard) product. The dimensions shown are for the ES policy (\Cref{table_arch}).}
\label{fig:gated_arch}
\vspace{-6mm}
\end{figure}

We represent a policy using a three-layer convolutional neural network (CNN) 
with gated activation units \cite{PixelCNN} (\Cref{fig:gated_arch}, \Cref{table_arch}). We found the gating mechanism was important for accelerating training and varied the number of channels depending on the algorithm (ES, PPO) for best performance.
\begin{table}
\centering
 \begin{tabular}{||l c c||} 
 \hline
 Policy type & ES & PPO\footnotemark \\
  \hline
 \hline
 Layers & 3 & 3 \\
 \hline
 Channels (per layer) & 8, 12, 8 & 8-32, 12-64, 8-32\\
 \hline
 Stride & 1,1,1  & 1,1,1   \\ 
 \hline
 Dilation & 1,2,4  & 1,2,4 \\
 \hline
 Activation function & $\tanh$  & $\tanh$  \\
 \hline
 Gating &  Y,Y,N  &  Y,Y,Y \\
 \hline
 Total parameters & 1.0k & 2.4k - 36k \\ [1ex] 
 \hline
\end{tabular}
\caption{Policy model details.}
\label{table_arch}
\vspace{-11mm}
\end{table}
\footnotetext{Architecture for both the value and policy networks. Total parameters are the sum of both networks' parameters.}

\Cref{fig:gated_arch} depicts this architecture. The input to the CNN is a tensor of shape $(T, S + 3)$, where $T$ is the number of past observations, and $S$ is the DOF of the robot. The $+ 3$ corresponds to a measurement of the 3D position of the ball, which is appended to the robot state. In our experiments, $S = 8$ and $T = 8$, so the input has shape $(8,11)$. The CNN applies 1D convolutions on the time dimension, with the ball and joint state treated as channels. Each gated layer produces two tensors $o_1$ (red) and $o_2$ (yellow) of equal size. The gating mechanism then multiplies the activations $\tanh(o_1)$ elementwise with the mask $\sigma(o_2)$ to produce the output $y_i$. Formally, let $W_i$ be the kernel of hidden layer $i$, $b_i$ the bias, and $X_i$ the input. We have
\vspace{-1mm}
\begin{align}
\begin{split}
(o_1, o_2) &= X_i \circledast W_i + b_i \\
y_i &= \tanh(o_1) \odot \sigma(o_2).
\vspace{-3mm}
\end{split}
\end{align}

To better understand the effect of using a CNN controller, we also trained simple three-layer multi-layer perceptron (MLP) controllers with 50 and 10 units in two hidden layers. 

\footnotetext{Drawn in PlotNeuralNet. \url{https://github.com/HarisIqbal88/PlotNeuralNet}}

\subsection{Curriculum Learning in RL}\label{sec:methods_curriculum}
The task of returning the ball defines a sparse reward of $1.0$ given at the end of episodes (ball throws) in which the robot succeeded, and $0.0$ otherwise. In principle, this type of sparse reward can be used to find an optimal policy as it exactly expresses the goal of the task. However, simply maximizing the success rate has two major drawbacks:
\begin{itemize}
\item Other considerations such as smoothness and style (see \Cref{sec:rtt_background}) are not captured by the sparse reward.
\item Sparse rewards make training difficult, as $\nabla F(\theta) = 0$ for a large portion of the parameter space. A large amount of exploration is required to observe any signal when rewards are highly sparse, which makes gradient estimation difficult. Furthermore, gradient-based algorithms are more likely to get trapped in \emph{local maxima}.
\end{itemize}

We can address these issues using \emph{curriculum learning} \cite{BLCW2009ICML} by carefully adjusting two aspects of the problem:

\begin{itemize}
\item Shaping the \emph{training distribution}, e.g. changing the distribution of ball throws during training so that a policy can improve on its weaknesses. In the MDP formalism for RL, this is changing the \emph{distribution of initial states} $s_{0} \sim \mathcal{P}(\mathcal{S})$.
\item Shaping the \emph{reward function}. New rewards can be added to e.g. discourage moving at high speeds, improve the swing pose, or to use forehand and backhand swings.
\end{itemize}

Curriculum learning is especially important for learning complex styles such as bimodal play, as we explore in \Cref{sec:playstyles}.

We briefly discuss an alternative to curriculum learning: human engineering of the desired behavior. Indeed, we can create a hierarchical policy which uses a fixed decision rule to select between sub-policies based on the predicted ball landing position. This is able to achieve a near-perfect success rate of 94\% (\Cref{tab:top_bimodal}, \emph{Hierarchical}) by leveraging optimal forehand and backhand policies trained with RL. However this approach is ultimately limited. Though rules-based engineering is possible for achieving the narrow goal of selecting between forehand-backhand, it is highly likely that more complex aspects of style or strategy cannot be neatly decomposed into distinct modes. Moreover, training separate policies for each mode quickly increases the total number of parameters. This limits the viability of engineering (it is not \emph{extensible}), and motivates our use of curriculum learning.

\subsection{Algorithms}

We consider two classes of RL optimization algorithms: \emph{evolutionary search} and \emph{policy gradient} methods.

\subsubsection{Evolutionary Search (ES)}

This class of optimization algorithms \cite{WSPS2008EC,NS2017FOCM} has recently become popular for RL \cite{SHCSS2017, stockholm, asebo, rbo}. ES is a general blackbox optimization paradigm, particularly efficient on objectives $F(\theta)$ which are possibly non-smooth. The key idea behind ES is to consider the \emph{Gaussian smoothing} of $F$, given by the following equation:
\vspace{-1mm}
\begin{equation}
F_{\sigma}(\theta) = \mathbb{E}_{\mathbf{g} \sim \mathcal{N}(0,\mathbf{I}_{d})}[F(\theta+\sigma \mathbf{g})],
\end{equation}
where $\sigma > 0$ controls the precision of the smoothing. It can be shown that $F_\sigma(\theta)$ is differentiable with gradients:
\begin{equation}
\label{es_grad}
\nabla F_{\sigma}(\theta) = \frac{1}{\sigma} \mathbb{E}_{\mathbf{g} \sim \mathcal{N}(0,\mathbf{I}_{d})} [F(\theta+\sigma \mathbf{g})\mathbf{g}],
\end{equation}
for which it is easy to derive unbiased Monte Carlo (MC) estimators. 
The ES method applies stochastic gradient ascent (SGD) to $F_\sigma(\theta)$, using the estimator $\widehat{\nabla} F_{\sigma}(\theta)$, so the update takes the form $\theta \leftarrow \theta + \eta \cdot \widehat{\nabla} F_{\sigma}(\theta)$. 

ES algorithms come in different variations, where the differentiation is made based on: Monte Carlo estimators used, additional performed normalizations, specific forms of control variate terms and more. We use ES methods with state normalization \cite{nagabandi,SHCSS2017}, filtering \cite{SHCSS2017}, and reward normalization \cite{MGR2018}, with repeated rollouts for each parameter perturbation $\mathbf{g_i}$. We found that averaging the reward from repeated rollouts was important for training good policies. This is likely due to the high degree of random variation between episodes. We also observed that the state-normalization heuristic was crucial for achieving good performance.

\subsubsection{Policy Gradient Methods}
Policy gradient (PG) methods
\cite{W1992ML,SMSM2000NIPS,SB1998MIT} are commonly used in RL and have been adapted for robotics \cite{LHPHETSW2016ICLR}. We experiment with a state-of-the-art PG algorithm called \emph{proximal policy optimization} (PPO) \cite{SWDRK2017arxiv}. Note that standard PG methods require \emph{stochastic} policies which is not the case for ES algorithms.

\section{Experiments}\label{sec:experiments}

Observations, results and conclusions from all conducted experiments are organized as follows:
\begin{itemize}
\item ES outperforms PPO in final reward, produces smoother policies, and requires fewer network parameters. (\Cref{sec:model-free-policies}).
\item CNN-based architectures both outperform MLP-based architectures and produce significantly smoother motions (\Cref{sec:architecture-results}).
\item Reward shaping can improve the policy's smoothness without impacting the success rate (\Cref{sec:unimodal_reward_shaping}).
\item Bimodal play is nontrivial to learn, but can be obtained by using curriculum learning (\Cref{sec:playstyles,sec:final_curriculum}). It does not require any modifications of the architecture.
\end{itemize}

\subsection{System Description}\label{sec:system_desc}

\begin{figure}
\centering
\includegraphics[width=0.45\textwidth]{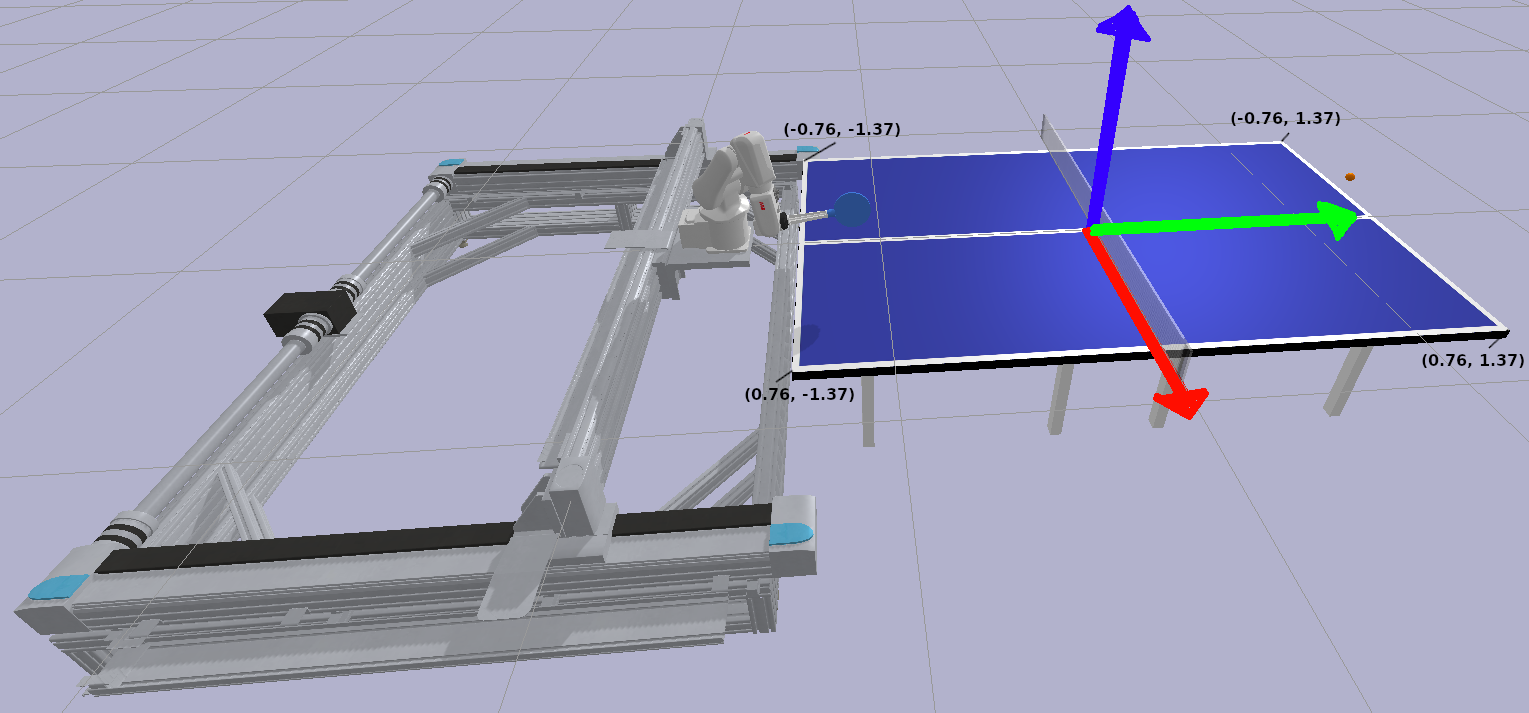}
\caption{Simulated robotic table tennis system. Our coordinate system places $(0,0,0)$ at the table center, and the axes are color-coded as $x=$ \textcolor{red}{red}, $y=$ \textcolor{green}{green}, $z=$ \textcolor{blue}{blue}.}
\label{robot_system}
\vspace{-6mm}
\end{figure}

Our robotic table tennis system consists of a 6DOF ABB IRB120 arm with revolute joints, and two linear axes permitting movement across and behind the table, for a total of 8DOF. The table conforms to ITTF standards and is 2.74m long, 1.525m wide, and the net is 15.25cm high. See \Cref{robot_system} for a depiction of the robot and table. The simulation is built using PyBullet \cite{coumans2019}.

Each episode consists of one ball throw, and the outcome is a \emph{hit} if the paddle makes contact with the ball, and a \emph{success} if it is hit and lands on the opponent's table. The simulation uses a simplified ball dynamics model that excludes air drag and spin. Ball throws are generated by randomly sampling the initial ball position\footnote{The coordinate system places $(0,0,0)$ at the table center, and the $x$ and $y$ axes are parallel to the width and length of the table. See \Cref{robot_system}.} $(x_0, y_0, z_0)$, the target landing coordinates on the robot side of the table $(x_1, y_1, 0)$, and the initial z-axis speed $v_z$. The full initial velocity vector $(v_x, v_y, v_z)$ is then solved for, and we accept throws with $v_y \in \lbrack -8.5, -3.5\rbrack$. Using this approach, we define two ball distributions: a \textit{forehand ball distribution} with $x_1 \in [-0.2, 0.7]$ and a \textit{full table distribution} with $x_1 \in [-0.7, 0.7]$. At the start of each ball throw, the arm is initialized to either a forehand pose (see \Cref{fig:poses}, LHS) or a central pose (see: \Cref{robot_system}) depending on the ball distribution. The initial pose is perturbed slightly to prevent overfitting.
 
For policy evaluation, we simulate 2500 episodes, and report success and hit rates, as well as smoothness metrics.

Randomness can be injected into the simulation in a number of different ways. Uniform random noise is added to the ball position at each timestep. Additionally, the ball and robot observations can be delayed independently and by a random number of simulation time steps each episode. Policy actions can also be delayed by a random number of time steps each episode.

\subsection{ES and PPO}\label{sec:model-free-policies}
 
We first trained CNN-based ES and PPO policies on the forehand ball distribution with sparse rewards: +1 for hitting the ball, +1 for landing the ball. ES policies were each trained for $15$K parameter updates, equivalently $22.7$M episodes, where as the PPO policies were trained for $2$M parameter updates, equivalently $1.1-1.2$M episodes.

ES policies were stronger in this setting, with 89\% successful returns to the opponent side of the table (see Table \ref{table:ppo-es-med}, Column S, ``sparse" suffix). PPO policies attained only 70\% success rate and required more parameters to train well on this task. For example, we found it was necessary to increase the number of parameters to $\sim 36$K to achieve the best performance. PPO policies with a comparable number of parameters ($\sim 2.4$K PPO vs $\sim 1.0$K ES) only has a success rate of 22\%. The phenomenon of policy gradient requiring more parameters than ES is not new, however. This effect has been observed numerous times in previous work. See for example: low-displacement rank policies with linear number of parameters or linear policies trained with ES outperforming PPO-trained ones 
\cite{stockholm, MGR2018}, as well as reverse trends between performance and architecture size for meta-learning (MAML) \cite{Song2020,FinnL18}.
Several explanations have been proposed previously for these effects, such as fewer parameters producing more stable and generalizable behaviors \cite{MGR2018, simplicity}. Meanwhile, for policy gradient methods, complex architectural requirements such as requiring stronger representational power for value approximation \cite{FinnL18}, or additional batch normalization layers \cite{LHPHETSW2016ICLR}, can be bottlenecks.

\begin{table}[ht]
\centering
 \begin{tabular}{||l c c c c c c||} 
 \hline
 Policy & S & H & J & A & V & JR \\ [0.5ex] 
 \hline\hline
  Random-8-12-8 & 0 & 2 & 1.5 & 1.0 & 0.9 & 6.1\\ 
 \hline
 ES-8-12-8-sparse & 89 & 99 & 4.0 & 2.9 & 3.4 & 9.5\\ 
 \hline
 PPO-8-12-8-sparse & 22 & 98 & 8.1 & 5.3 & 4.5 & 14.1\\ 
 \hline
 PPO-16-32-16-sparse & 70 & 98 & 9.6 & 6.2 & 4.6 & 14.2\\ 
 \hline
 PPO-32-64-32-sparse & 70 & 97 & 8.9 & 5.8 & 4.4 & 14.3\\ 
 \hline \hline
 ES-8-12-8-shaped & \textbf{90} & 99 & 2.9 & 2.2 & 2.9 & 9.2 \\ 
 \hline
 ES-8-12-8-shaped + AF 3Hz & 79 & 98 & \textbf{2.0} & \textbf{1.4} & \textbf{2.1} & \textbf{9.0} \\ 
 \hline
 PPO-8-12-8-shaped & 33 & 93 & 5.9 & 3.9 & 3.9 & 12.2 \\ 
 \hline
 PPO-16-32-16-shaped & 50 & 98 & 4.4 & 3.2 & 4.0 & 11.6 \\ 
 \hline
 PPO-32-64-32-shaped & 62 & 99 & 4.1 & 3.3 & 3.9 & 11.5\\  [1ex]
 \hline
\end{tabular}
\caption{Results: Forehand ball distribution S: success (\%), H: hit (\%), J: avg max jerk, A: avg max acceleration, V: avg max velocity, JR: sum of joint range.\protect\footnotemark}
\label{table:ppo-es-med}
\vspace{-6mm}
\end{table}
\footnotetext{Bold indicates best performance in a column, excluding the random policy.}

We observe a similar performance differential between ES and PPO for the harder task of the full table ball distribution as shown in Table \ref{table:ppo-es-hard}. For this reason we conducted the ablation studies and training for bimodal play using only ES.

\addtocounter{footnote}{-1}
\begin{table}[ht]
\centering
 \begin{tabular}{||l c c c c c c||} 
 \hline
 Policy & S & H & J & A & V & JR \\ [0.5ex] 
 \hline\hline
 Random-8-12-8 & 0 & 1 & 0.7 & 0.5 & 0.8 & 8.6\\
 \hline
 ES-8-12-8-sparse & 39 & 97 & 5.5 & 3.6 & 3.6 & 10.9 \\
 \hline
 PPO-8-12-8-sparse & 18 & 89 & 6.3 & 4.5 & 3.7 & 11.1 \\ 
 \hline
 PPO-16-32-16-sparse & 21 & 94 & 7.0 & 4.9 & 4.4 & 14.5 \\ 
 \hline
 PPO-32-64-32-sparse & 34 & 95 & 9.2 & 6.0 & 4.4 & 14.4 \\ 
 \hline \hline
 ES-8-12-8-shaped & \textbf{48} & 97 & 4.6 & 3.1 & 3.7 & 11.8 \\
 \hline
 ES-8-12-8-shaped + AF 3Hz & 8 & 98 & 2.2 & 2.0 & 2.7 & 11.8 \\ 
 \hline
 PPO-8-12-8-shaped & 1 & 52 & \textbf{2.0} & \textbf{1.6} & \textbf{2.4} & \textbf{10.7} \\ 
 \hline
 PPO-16-32-16-shaped & 4 & \textbf{98} & 4.1 & 3.3 & 4.0 & 14.4 \\ 
 \hline
 PPO-32-64-32-shaped & 31 & 96 & 4.2 & 3.9 & 4.2 & 13.6 \\  [1ex]
 \hline
\end{tabular}
\caption{Results: Full table ball distribution. S: success (\%), H: hit (\%), J: avg max jerk, A: avg max acceleration, V: avg max velocity, JR: sum of joint range.\protect\footnotemark}
\label{table:ppo-es-hard}
\vspace{-11mm}
\end{table}

\subsection{Policy architecture}\label{sec:architecture-results} 
Table \ref{table_mlp_cnn} shows that policy network architecture has a significant effect on both success rates and smoothness. We evaluate gated CNN and MLP policies, each with 3 hidden layers. Both policy types received the same inputs: the 8 most recent time-steps of ball and joint positions.

To evaluate smoothness we look at three metrics averaged over all time steps and joints: (1) maximum jerk per time step (J), (2) maximum acceleration per time step (A), and (3) maximum velocity per time step (V). We also measure the joint range (JR) of a policy. 

CNN policies achieved higher success rates in both the sparse reward and shaped reward (see \ref{sec:unimodal_reward_shaping}) settings compared to MLPs: 89\% vs. 67\% with sparse reward and 90\% vs 46\% with shaped rewards. MLP policies are significantly less smooth, with average max. jerk of ~2.5 - 3x that of the CNN, 2.5x average max. acceleration, and 1.3 - 1.5x average max. velocity. MLPs also have a noticeably higher range of motion when compared with CNNs (column JR, Table \ref{table_mlp_cnn}). These differences are also clear from a qualitative inspection of the policy behavior (see policies B and C in the supplementary video).

\begin{table}[ht]
\centering
 \begin{tabular}{||l c c c c c c||} 
 \hline
 Architecture & S & H & J & A & V & JR \\ [0.5ex] 
 \hline\hline
 CNN (sparse) & 89 & \textbf{99} & 4.0 & 2.9 & 3.4 & 9.5   \\ 
 \hline
 MLP (sparse) & 67 & 98 & 11.1 & 7.2 & 4.4 & 11.4  \\
 \hline
 CNN (shaped) & \textbf{90} & \textbf{99} & \textbf{2.9} & \textbf{2.2} & \textbf{2.9} & \textbf{9.2}   \\ 
 \hline
 MLP (shaped) & 46 & 95 & 7.6 & 5.4 & 4.3 & 13.4 \\ [1ex]
 \hline
\end{tabular}
\caption{Comparing CNN and MLP policies with sparse and shaped rewards on the forehand ball distribution. All policies were trained for 15K parameter updates.}
\label{table_mlp_cnn}
\vspace{-8mm}
\end{table}

\subsection{Reward shaping}\label{sec:unimodal_reward_shaping}
\addtocounter{footnote}{1}
Reward shaping is common practice in RL and it is an effective technique for shaping policy behavior. We explored the effect of seven different rewards (see footnote \thefootnote) shown in Table \ref{table:reward_shaping}, and find that it improves style with little or no cost in success rate for the same number of optimization steps.
\addtocounter{footnote}{-1}
\begin{table}[ht]
\centering
 \begin{tabular}{||l c c c c c c||} 
 \hline
 Rewards\footnotemark & S & H & J & A & V & JR \\ [0.5ex] 
 \hline\hline
  ST (sparse) & 89  & \textbf{99} & 4.0 & 2.9 & 3.4  & 9.5 \\ 
 \hline
  ST, IC, BBR & 87 & 99 & 4.3 & 3.1 & 3.1 & 8.6 \\ 
 \hline
  ST, IC, BBR, PH, JA  & \textbf{91} & \textbf{99} & 3.2 & 2.5 & 3.1 & 9.7 \\ 
 \hline
  ST, IC, BBR, PH, V, A, J & 96 & \textbf{99} & \textbf{2.7} & \textbf{2.0} & \textbf{2.8} & \textbf{8.5} \\ 
 \hline
  Canonical (all rewards)\footnotemark & 90 & \textbf{99} & 2.9 & 2.2 & 2.9 & 9.2 \\  [1ex]
 \hline
\end{tabular}
\caption{Ablation study: Effect of different rewards on success, hit rates, and smoothness metrics. Policies were trained and evaluated on the forehand ball distribution.}
\label{table:reward_shaping}
\vspace{-8mm}
\end{table}
\addtocounter{footnote}{-1}
\footnotetext{ST: hit and success sparse rewards, IC: penalty for self-collision or colliding with the table, BBR: penalty for rotating the base joint too far backwards, PH: penalty if the paddle gets too close to the table, JA: penalty if the arm position gets too close to any of the joint limits, V / A / JL: Penalty for exceeding a velocity / acceleration / jerk limit.}
\stepcounter{footnote}
\footnotetext{Canonical reward shaping: ST, IC, BBR, PH, V, A, J, JA}

We observe a similar pattern with PPO in that reward shaping improves the smoothness metrics (see \Cref{table:ppo-es-med} for example). However, PPO policies also are noticeably less smooth, with J, A, and V values over 2x that of comparable ES policies, and have a larger range of motion. It may be that PPO results in more sensitive and mobile policies compared to ES, and this makes it harder to train smooth policies. It would be interesting to explore this further in future work.

\subsection{Action filters}\label{sec:action-filter}
We find that applying a low-pass butterworth filter to the policy actions further increases smoothness (see Table \ref{table_action_filter}) but the differences in smoothness metrics between filters with varying cutoff frequencies is not very large. Unlike reward shaping, adding a filter appears to make the primary problem of returning balls harder, and leads to slightly lower success rates. 

\begin{table}[ht]
\centering
 \begin{tabular}{||l c c c c c c||} 
 \hline
 Action filter & S & H & J & A & V & JR \\ [0.5ex] 
 \hline\hline
  None  & \textbf{90} & \textbf{99} & 2.9 & \textbf{2.1} & 2.9 & 9.2\\ 
 \hline
  f-cut: 2Hz & 84 & 97 & \textbf{1.8} & \textbf{1.2} & 2.2 & 9.0 \\ 
 \hline
  f-cut: 3Hz & 79 & 98 & 2.0 & 1.4 & \textbf{2.1} & 9.0   \\ 
 \hline
  f-cut: 5Hz & 80 & \textbf{99} & 1.9 & 1.3 & 2.2 & \textbf{7.2} \\ 
 \hline
  f-cut: 10Hz & 88 & \textbf{99} & 1.9 & 1.5 & 2.4 & 9.5   \\  [1ex]
 \hline
\end{tabular}
\caption{Ablation study: Effect of applying a low-pass action filter to the policy output. Policies were trained and evaluated on the forehand ball distribution.}
\label{table_action_filter}
\vspace{-8mm}
\end{table}

\subsection{Learning Complex Playstyles}\label{sec:playstyles}
Policy A, as shown in the video, is the result of training with a curriculum on the ball distribution and rewards. It has a success rate of 80\% on the full table ball distribution, a hit rate of 99\%, and has a good bimodal style, with balanced success rates on the forehand and backhand (\Cref{tab:top_bimodal}). We also note that Policy A can obtain success rates nearing the human-engineered hierarchical policy (discussed in \Cref{sec:methods_curriculum}) which can be seen as a near-optimal policy for this problem, despite Policy A having a much smaller model.

\begin{table}[h]
    \centering
    \begin{tabular}{||ccccccc||}
    \hline
    Policy & S\footnotemark & H & S-F & H-F & S-B & H-B \\ \hline \hline
    \emph{A} & 80 & 99 & 78 & 99 & 81 & 99 \\ \hline
    Hierarchical & 94 & 100 & 92 & 100 & 96 & 100 \\ \hline
    \end{tabular}
    \caption{Performance of bimodal policies on the full table ball distribution. Policy A is shown in video.}
    \label{tab:top_bimodal}
    \vspace{-6mm}
\end{table}

\footnotetext{ `S' denotes success rate, `H' denotes hit rate, `F' denotes the rate for ball throws on the forehand side of the table, and `B' the backhand.} 

The curriculum used to train Policy A is listed in \Cref{sec:final_curriculum}. To understand the necessity of each component, we carry out a series of ablation studies, shown in \Cref{tab:bimodal_ablations}. In each ablation test, we train a policy for 15K steps and evaluate its success rate and style. Note that success rates are low overall, since training with the \emph{full table distribution} (\Cref{sec:system_desc}) requires more steps than the \emph{forehand table distribution} used in the ablation studies of \Cref{sec:model-free-policies}.

A major challenge is that optimizing for success rates typically leads to a unimodal policy. For instance, in \Cref{tab:bimodal_ablations}, row \#1 shows that under the canonical rewards, the resulting policy is forehand-only; it achieves 23\% success on forehand balls but only 0.8\% on backhand balls. This is also confirmed by visual inspection. To reliably obtain bimodal play, we must devise a curriculum to encourage it. First, we introduce \emph{pose rewards} which can maintain bimodal style (Section \ref{sec:bimodal_pose_reward}). Then we discuss how adding success bonuses (\Cref{sec:landing_bonus}) and shaping the task distribution can improve training (Section \ref{sec:task_shaping}). Finally in Section \ref{sec:final_curriculum} we present the curriculum for training Policy A.

\addtocounter{footnote}{-1}
\begin{table}[t]
    \centering
    \scalebox{0.93}{
    \begin{tabular}{||c|c c c c c c c||}
    \hline 
         \# & Regime & S\footnotemark & H & S-F & H-F & S-B & H-B \\ \hline \hline
          & Pose Reward & \multicolumn{6}{l|}{(\emph{see \Cref{sec:bimodal_pose_reward}})} \\ \hline
        1 & None & 11.8 & 88.3 & 23.6 & 96.3 & 0.8 & 80.8 \\ \hline
        2 & CPS & 17.7 & 53.3 & 28.4 & 73.8 & 6.3 & 31.5 \\ \hline
        \textbf{3} & \textbf{DCPS} & 2.3 & 83.7 & 1.4 & 85.4 & 3.3 & 82.1 \\ \hline
        \textbf{4} & \textbf{CPT} & 0.8 & 95.3 & 1.3 & 95.2 & 0.4 & 95.5 \\ \hline
        & Success + Pose & \multicolumn{6}{l|}{(\emph{see \Cref{sec:landing_bonus}})} \\ \hline
        5 & DTR + None & 9.6 & 71.4 & 16.8 & 92.4 & 2.4 & 50.6 \\ \hline
        6 & DTR + DCPS & 11.3 & 88.3 & 18.9 & 98.0 & 2.7 & 77.3 \\ \hline
        & Ball Range + Pose & \multicolumn{6}{l|}{(\emph{see \Cref{sec:task_shaping}})} \\ \hline
        7 & $(0.5,0.7)$ + None & 16.4 & 36.9 & 32.6 & 73.6 & 0 & 0 \\ \hline
        8 & $(0.3, 0.7)$ + None & 0.3 & 38.3 & 0.6 & 77.0 & 0 & 0 \\ \hline
        9 & $(0.3, 0.7)$ + CPS & 3.6 & 19.1 & 7.0 & 37.5 & 0 & 0 \\ \hline
        \textbf{10} & \textbf{ $(0.3, 0.7)$ + DCPS} & 8.5 & 87.6 & 9.7 & 87.4 & 7.3 & 87.5 \\ \hline
    \end{tabular}}
    \caption{Ablation studies: performance after 15K steps for various regimes. Numbers express percentages, and bolded rows (\textbf{3,4, 10}) indicate bimodal policies (from visual inspection).}
    \label{tab:bimodal_ablations}
    \vspace{-11mm}
\end{table}

\subsubsection{Reward Shaping for Bimodal Play}\label{sec:bimodal_pose_reward}
We have already seen reward shaping in \Cref{sec:unimodal_reward_shaping} to encourage better style. We use a similar technique for bimodal play.

\begin{figure}
\centering
\includegraphics[width=0.24\textwidth]{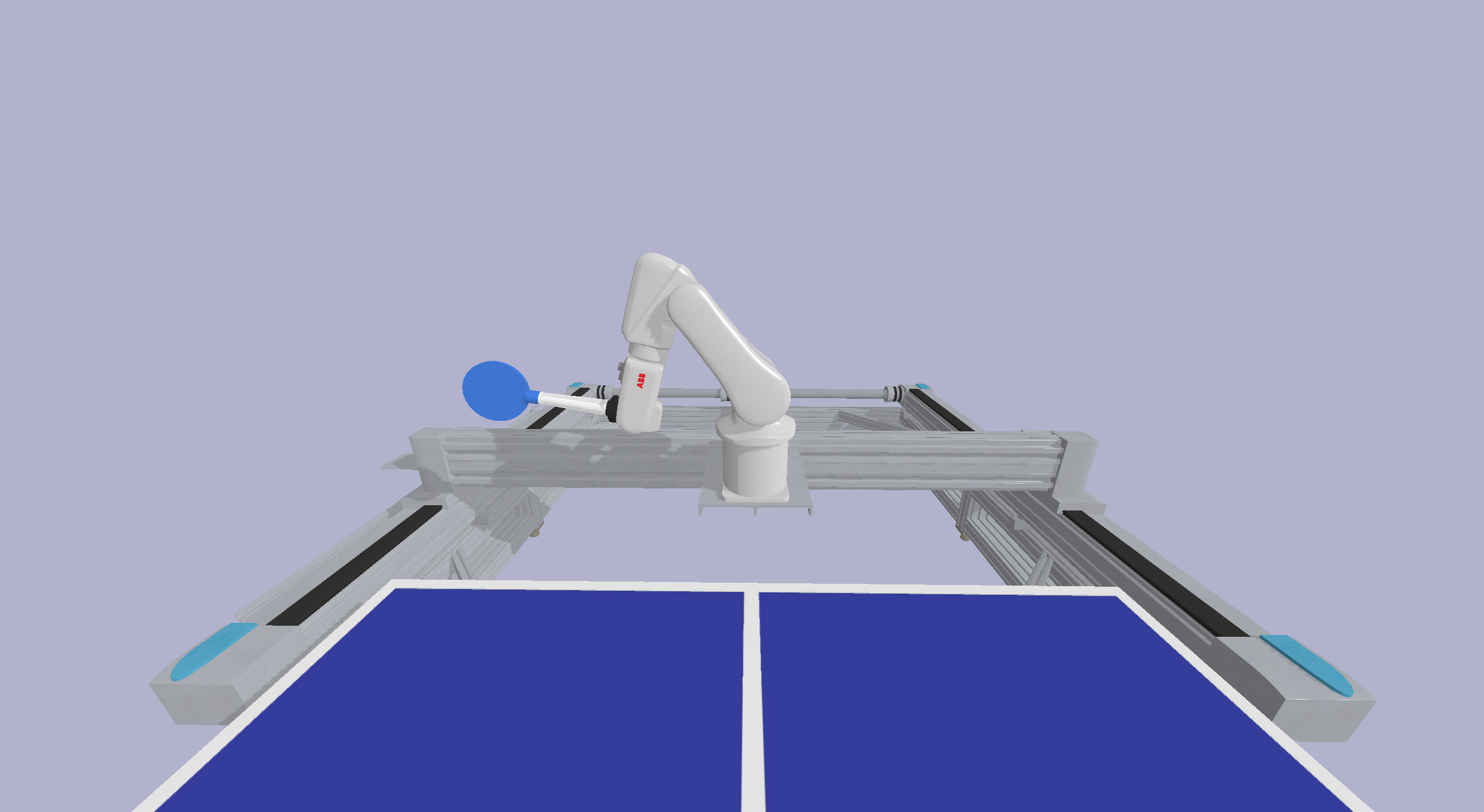}
\includegraphics[width=0.24\textwidth]{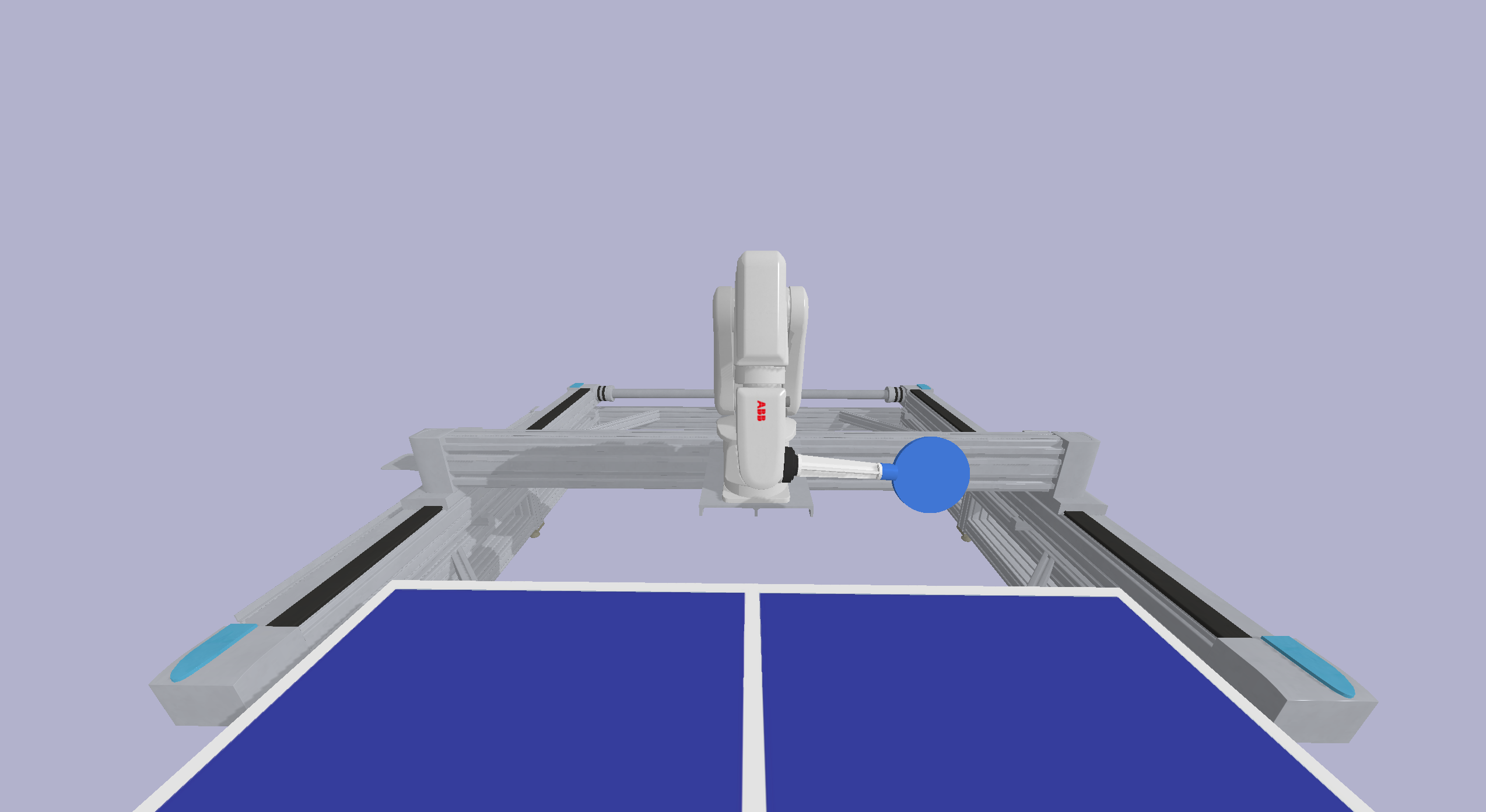}
\caption{Reference forehand (left) and backhand (right) poses.}
\label{fig:poses}
\vspace{-5mm}
\end{figure}

Precisely characterizing `forehand' swings and the ball throws for which a forehand swing should be rewarded, is nontrivial, and vary in the amount of human knowledge implicit in the reward definition. We consider several variants of pose rewards:
\begin{enumerate}
\item \textbf{Conditional Pose State} (CPS): We define reference poses for the forehand and backhand, which are shown in \Cref{fig:poses}. The reward is given for taking a pose close\footnote{As measured by the $L_2$ norm in joint space} to the reference pose corresponding to the side on which the ball lands. That is, $R^{F}_{CPS} = 1 - d^F$ is awarded for episodes where the ball lands on the forehand side, where $d^F$ is the minimum distance of the arm's pose to the forehand reference point, and $R^{B}_{CPS}$ is defined similarly for backhand.
\item \textbf{Dense Conditional Pose State} (DCPS): A denser version of the CPS reward which penalizes taking the wrong pose. The reward is $w(R^{F}_{CPS} - R^{B}_{CPS})$ for episodes where the ball is thrown to the forehand, and $w(R^{B}_{CPS} - R^{F}_{CPS})$ for throws to the backhand. We add a scale $w$ which reduces the magnitude of the reward if the ball's landing point is near the center.
\item \textbf{Conditional Pose Timesteps} (CPT): We define forehand and backhand in terms of the rotation of J1 and J4 (\Cref{robot_arm}). The pose is considered to be `forehand' if J1 and J4 both lie in the appropriate half of their ranges, and similarly for backhand. Let $t^F$ to be the percentage of timesteps before ball contact such that the robot is in a forehand pose, and similarly for $t^B$ on the backhand. The reward given is $R^F_{CPT} = w(t^F - t^B)$ for balls thrown to the forehand, and $R^B_{CPT} = w(t^B - t^F)$ for backhand balls.
\end{enumerate}

We first investigate each type of pose reward (rows \#1-4, \Cref{tab:bimodal_ablations}). As noted before, training without pose rewards led to a purely forehand policy (row \#1). The CPS reward also led to a forehand policy (row \#2), indicating that locally, the policy can still obtain higher reward by improving its forehand play at the expense of the backhand. This was a motivation for the DCPS reward (row \#3). Although its success rate after the same period of training is lower, its hit rate is high, and the forehand and backhand success rates are balanced. Video inspection confirms it has a basic bimodal style. The CPT reward produces similar training as DCPS in early stages, but video inspection shows that DCPS produces a clearer bimodal style. In general, a more `prescriptive' reward such as DCPS appears to speed up learning at the beginning (especially with distribution shaping; see \Cref{sec:task_shaping}), whereas a more `flexible' reward such as CPT has advantages in later training (\Cref{sec:landing_bonus}).

\subsubsection{Reward Shaping to Escape Plateaus}\label{sec:landing_bonus}
Bimodal policies often reach a plateau in training where further progress becomes extremely slow. This is likely because of the sparsity of the success reward; since it does not distinguish between return balls that are close to landing and those that are far from the target, it is difficult for exploration to find nearby policies with sufficiently higher success rates so as to have measurably better rewards. We experiment with two rewards for increasing the success signal:
\begin{enumerate}
\item \textbf{Landing Bonus}: Increase the sparse success reward.
\item \textbf{Distance to Table} (DTR): A dense version of the success reward given by $\max\{1 - d, -2\}$, where $d$ is the minimum distance between the ball and the opponent's table surface during the return trajectory.
\end{enumerate}

Introducing these rewards helps to escape plateaus for policies that are \emph{already} bimodal and have reasonable success rates. Training such policies with the CPT pose reward and the DTR success reward leads to continued improvement while maintaining style, especially combined with distribution shaping (\Cref{sec:task_shaping}). However, we find that the balance between success rewards and pose rewards which lead to bimodal play can be sensitive. For example;
\begin{enumerate}
    \item Using DTR at the onset leads to a unimodal policy, even with DCPS pose rewards (rows \#5, 6). 
    \item When initialized with a bimodal policy (trained with DCPS), subsequent training with DTR and DCPS leads the policy to collapse to an unusual unimodal style which takes a backhand pose and then manipulates its `shoulder' J3 (video, Policy E) to reduce penalties.
    \item The combination of DCPS and a landing bonus of $+1.0$ leads to collapse. 
\end{enumerate}

\subsubsection{Shaping the Training Distribution}\label{sec:task_shaping} Varying the distribution of tasks can improve training. For instance, if our objective is forehand play only, notice by comparing \Cref{table:reward_shaping} to \Cref{tab:bimodal_ablations} that much higher success rates are obtained in 15K steps by restricting the ball distribution to forehand-only. Adjusting the difficulty of the task can therefore make training faster. It may also help to avoid local minima for which undesirable style is compensated by locally maximal success rates.

As we might expect, our experiments show that the policy has the greatest difficulty learning a bimodal style for balls that land at the center of the table. Based on this observation, we consider a spectrum of tasks where the ball's training distribution is supported on the two sides of the table. The \emph{ball range} $(a,b)$ for $0 \leq a < b$ indicates that the $x$-coordinate of the ball's landing position belongs to $\lbrack -b,-a\rbrack \cup \lbrack a,b \rbrack$. To overcome the style collapse problem when training a bimodal policy with the DTR reward, we change the ball range to $(0.5,0.7)$ when the DTR reward is introduced, and then periodically widen the range as the policy improves until the distribution spans the entire table.

We also consider whether this distribution shaping can be applied at the onset. However, ablation studies indicate that the policy learns to ignore one side of the table entirely (see rows \#7-9 of \Cref{tab:bimodal_ablations}), unless the DCPS reward is also given, which trains well. This shows that training is most efficient with a curriculum, which leads the policy to improve skills without forgetting previous ones.

\subsection{Effective Curriculum Learning}\label{sec:final_curriculum}

Based on the observations of the preceding sections, we used the following scheme to train bimodal policies with RL, an example of which is Policy A.
\begin{enumerate}
\item Canonical style rewards and DCPS pose reward on the full table.
\item Introduce the DTR success reward, switch the pose reward to CPT, and set the ball range to $(0.5, 0.7)$.
\item Increase the ball range to $(0.3,0.7)$, $(0.1,0.7)$, and then to the full table.
\end{enumerate}
The precise number of steps required varies owing to the randomness in RL algorithms; the first stage may take 15K steps, as was used in our ablation studies.

\section{Conclusions}
We have shown that model-free reinforcement learning is an effective approach for robotic table tennis and is suitable for high-speed control, generating actions at 100Hz. This has the advantage of avoiding ball prediction modeling and trajectory optimization without the need for human demonstrations. Policies can learn to hit and return balls simply by being given sparse rewards for contact and success. Moreover, reward shaping and curriculum learning can be used to improve the style of the policy, and develop more complex play. We demonstrated this by training strong bimodal policies which are capable of playing both the forehand and backhand. This suggests that RL is promising for robotic table tennis. Our experiments show that CNN policies outperform MLP policies both in terms of success and smoothness. We also observe that the evolution strategy (ES) paradigm in RL was particularly effective for this problem, and in comparison to policy gradient methods, was able to train smaller policies.

\section{Acknowledgements}

We thank David D'Ambrosio, Jie Tan, and Peng Xu for their helpful comments on this work.

\renewcommand*{\bibfont}{\footnotesize}
\bibliographystyle{plainnat}
\bibliography{root}

\begin{thebibliography}{47}
\providecommand{\natexlab}[1]{#1}
\providecommand{\url}[1]{\texttt{#1}}
\expandafter\ifx\csname urlstyle\endcsname\relax
  \providecommand{\doi}[1]{doi: #1}\else
  \providecommand{\doi}{doi: \begingroup \urlstyle{rm}\Url}\fi

\bibitem[Akrour et~al.(2016)Akrour, Abdolmaleki, Abdulsamad, Peters, and
  Neumann]{Akrour2016ModelFreeTP}
Riad Akrour, Abbas Abdolmaleki, Hany Abdulsamad, Jan Peters, and Gerhard
  Neumann.
\newblock Model-free trajectory-based policy optimization with monotonic
  improvement.
\newblock \emph{J. Mach. Learn. Res.}, 2016.

\bibitem[Anderson(1988)]{Anderson1988ARP}
Russell Anderson.
\newblock \emph{A Robot Ping-Pong Player: Experiments in Real-Time Intelligent
  Control}.
\newblock MIT Press, 1988.

\bibitem[Bengio et~al.(2009)Bengio, Louradour, Collobert, and
  Weston]{BLCW2009ICML}
Yoshua Bengio, J{\'e}r{\^o}me Louradour, Ronan Collobert, and Jason Weston.
\newblock Curriculum learning.
\newblock In \emph{ICML}, pages 41--48, 2009.

\bibitem[Billingsley(1983)]{Billingsley83}
John Billingsley.
\newblock Robot ping pong.
\newblock \emph{Practical Computing}, 1983.

\bibitem[Choromanski et~al.(2018)]{stockholm}
Krzysztof Choromanski et~al.
\newblock Structured evolution with compact architectures for scalable policy
  optimization.
\newblock In \emph{{ICML}}, 2018.

\bibitem[Choromanski et~al.(2019{\natexlab{a}})]{asebo}
Krzysztof Choromanski et~al.
\newblock From complexity to simplicity: Adaptive es-active subspaces for
  blackbox optimization.
\newblock In \emph{NeurIPS}, 2019{\natexlab{a}}.

\bibitem[Choromanski et~al.(2019{\natexlab{b}})]{rbo}
Krzysztof Choromanski et~al.
\newblock Provably robust blackbox optimization for reinforcement learning.
\newblock In \emph{CoRL}, 2019{\natexlab{b}}.

\bibitem[Coumans and Bai(2016--2019)]{coumans2019}
Erwin Coumans and Yunfei Bai.
\newblock Pybullet, a python module for physics simulation for games, robotics
  and machine learning.
\newblock \url{http://pybullet.org}, 2016--2019.

\bibitem[Finn and Levine(2018)]{FinnL18}
Chelsea Finn and Sergey Levine.
\newblock Meta-learning and universality: Deep representations and gradient
  descent can approximate any learning algorithm.
\newblock In \emph{ICLR}, 2018.

\bibitem[Gao et~al.(2019)Gao, Tebbe, Krismer, and Zell]{Gao2019MarkerlessRP}
Yapeng Gao, Jonas Tebbe, Julian Krismer, and Andreas Zell.
\newblock Markerless racket pose detection and stroke classification based on
  stereo vision for table tennis robots.
\newblock \emph{IEEE Robotic Computing}, 2019.

\bibitem[Hartley(1983)]{Hartley87}
J.~Hartley.
\newblock Toshiba progress towards sensory control in real time.
\newblock \emph{The Industrial Robot 14-1}, pages 50--52, 1983.

\bibitem[Hashimoto et~al.(1987)Hashimoto, Ozaki, and
  Osuka]{Hashimoto1987DevelopmentOP}
Hideaki Hashimoto, Fumio Ozaki, and Kuniji Osuka.
\newblock Development of ping-pong robot system using 7 degree of freedom
  direct drive robots.
\newblock In \emph{Industrial Applications of Robotics and Machine Vision},
  1987.

\bibitem[Huang et~al.(2015)Huang, Sch{\"o}lkopf, and
  Peters]{Huang2015LearningOS}
Yanlong Huang, Bernhard Sch{\"o}lkopf, and Jan Peters.
\newblock Learning optimal striking points for a ping-pong playing robot.
\newblock \emph{IROS}, 2015.

\bibitem[Huang et~al.(2016)Huang, Buchler, Koç, Sch{\"o}lkopf, and
  Peters]{Huang2016JointlyLT}
Yanlong Huang, Dieter Buchler, Okan Koç, Bernhard Sch{\"o}lkopf, and Jan
  Peters.
\newblock Jointly learning trajectory generation and hitting point prediction
  in robot table tennis.
\newblock \emph{IEEE-RAS Humanoids}, 2016.

\bibitem[Ijspeert et~al.(2002)Ijspeert, Nakanishi, and
  Schaal]{Ijspeert2002MovementIW}
Auke~Jan Ijspeert, Jun Nakanishi, and Stefan Schaal.
\newblock Movement imitation with nonlinear dynamical systems in humanoid
  robots.
\newblock \emph{ICRA}, 2002.

\bibitem[James et~al.(2019)James, Wohlhart, Kalakrishnan, Kalashnikov, Irpan,
  Ibarz, Levine, Hadsell, and Bousmalis]{james}
Stephen James, Paul Wohlhart, Mrinal Kalakrishnan, Dmitry Kalashnikov, Alex
  Irpan, Julian Ibarz, Sergey Levine, Raia Hadsell, and Konstantinos Bousmalis.
\newblock Sim-to-real via sim-to-sim: Data-efficient robotic grasping via
  randomized-to-canonical adaptation networks.
\newblock In \emph{CVPR}, 2019.

\bibitem[Kalashnikov et~al.(2018)]{QTOPT2018CORL}
Dmitry Kalashnikov et~al.
\newblock Qt-opt: Scalable deep reinforcement learning for vision-based robotic
  manipulation.
\newblock \emph{CoRL}, 2018.

\bibitem[Knight and Lowery(1986)]{Knight1986PingpongplayingRC}
John Knight and David Lowery.
\newblock Pingpong-playing robot controlled by a microcomputer.
\newblock \emph{Microprocessors and Microsystems - Embedded Hardware Design},
  1986.

\bibitem[Koç et~al.(2018)Koç, Maeda, and Peters]{Ko2018OnlineOT}
Okan Koç, Guilherme Maeda, and Jan Peters.
\newblock Online optimal trajectory generation for robot table tennis.
\newblock \emph{Robotics \& Autonomous Systems}, 2018.

\bibitem[Lillicrap et~al.(2016)Lillicrap, Hunt, Pritzel, Heess, Erez, Tassa,
  Silver, and Wierstra]{LHPHETSW2016ICLR}
Timothy Lillicrap, Jonathan Hunt, Alexander Pritzel, Nicolas Heess, Tom Erez,
  Yuval Tassa, David Silver, and Daan Wierstra.
\newblock Continuous control with deep reinforcement learning.
\newblock \emph{ICLR}, 2016.

\bibitem[Mahjourian et~al.(2018)Mahjourian, Jaitly, Lazic, Levine, and
  Miikkulainen]{Mahjourian2018HierarchicalPD}
Reza Mahjourian, Navdeep Jaitly, Nevena Lazic, Sergey Levine, and Risto
  Miikkulainen.
\newblock Hierarchical policy design for sample-efficient learning of robot
  table tennis through self-play.
\newblock \emph{arXiv:1811.12927}, 2018.

\bibitem[Mania et~al.(2018)Mania, Guy, and Recht]{MGR2018}
Horia Mania, Aurelia Guy, and Benjamin Recht.
\newblock Simple random search provides a competitive approach to reinforcement
  learning.
\newblock \emph{NeurIPS}, 2018.

\bibitem[Matsushima et~al.(2003)Matsushima, Hashimoto, and
  Miyazaki]{Matsushima2003LearningTT}
Michiya Matsushima, Takaaki Hashimoto, and Fumio Miyazaki.
\newblock Learning to the robot table tennis task-ball control and rally with a
  human.
\newblock \emph{IEEE International Conference on Systems, Man and Cybernetics},
  2003.

\bibitem[Matsushima et~al.(2005)Matsushima, Hashimoto, Takeuchi, and
  Miyazaki]{Matsushima2005ALA}
Michiya Matsushima, Takaaki Hashimoto, Masahiro Takeuchi, and Fumio Miyazaki.
\newblock A learning approach to robotic table tennis.
\newblock \emph{IEEE Transactions on Robotics}, 2005.

\bibitem[Miyazaki et~al.(2002)Miyazaki, Takeuchi, Matsushima, Kusano, and
  Hashimoto]{Miyazaki2002RealizationOT}
Fumio Miyazaki, Masahiro Takeuchi, Michiya Matsushima, Takamichi Kusano, and
  Takaaki Hashimoto.
\newblock Realization of the table tennis task based on virtual targets.
\newblock \emph{ICRA}, 2002.

\bibitem[Miyazaki et~al.(2006)]{Miyazaki2006LearningTD}
Fumio Miyazaki et~al.
\newblock Learning to dynamically manipulate: A table tennis robot controls a
  ball and rallies with a human being.
\newblock In \emph{Advances in Robot Control}, 2006.

\bibitem[Muelling and Peters(2009)]{Muelling2009ACM}
Katharina Muelling and Jan Peters.
\newblock A computational model of human table tennis for robot application.
\newblock In \emph{AMS}, 2009.

\bibitem[Muelling et~al.(2010{\natexlab{a}})Muelling, Kober, and
  Peters]{Muelling2010Biomem}
Katharina Muelling, Jens Kober, and Jan Peters.
\newblock A biomimetic approach to robot table tennis.
\newblock \emph{Adaptive Behavior}, 2010{\natexlab{a}}.

\bibitem[Muelling et~al.(2010{\natexlab{b}})Muelling, Kober, and
  Peters]{Muelling2010LearningTTMOMP}
Katharina Muelling, Jens Kober, and Jan Peters.
\newblock Learning table tennis with a mixture of motor primitives.
\newblock \emph{IEEE-RAS Humanoids}, 2010{\natexlab{b}}.

\bibitem[Muelling et~al.(2010{\natexlab{c}})Muelling, Kober, and
  Peters]{Muelling2010SimulatingHT}
Katharina Muelling, Jens Kober, and Jan Peters.
\newblock Simulating human table tennis with a biomimetic robot setup.
\newblock In \emph{Simulation of Adaptive Behavior}, 2010{\natexlab{c}}.

\bibitem[Muelling et~al.(2012)Muelling, Kober, Kroemer, and
  Peters]{Muelling2012LearningSelectGen}
Katharina Muelling, Jens Kober, Oliver Kroemer, and Jan Peters.
\newblock Learning to select and generalize striking movements in robot table
  tennis.
\newblock \emph{The International Journal of Robotics Research}, 2012.

\bibitem[Nagabandi et~al.(2018)]{nagabandi}
Anusha Nagabandi et~al.
\newblock Neural network dynamics for model-based deep reinforcement learning
  with model-free fine-tuning.
\newblock In \emph{ICRA}, 2018.

\bibitem[Nesterov and Spokoiny(2017)]{NS2017FOCM}
Yurii Nesterov and Vladimir Spokoiny.
\newblock Random gradient-free minimization of convex functions.
\newblock \emph{FoCM}, 2017.

\bibitem[Rajeswaran et~al.(2017)Rajeswaran, Lowrey, Todorov, and
  Kakade]{simplicity}
Aravind Rajeswaran, Kendall Lowrey, Emanuel Todorov, and Sham~M. Kakade.
\newblock Towards generalization and simplicity in continuous control.
\newblock In \emph{NeurIPS}, 2017.

\bibitem[Ramanantsoa and Durey(1994)]{Ramanantsoa94}
Marie-Marin Ramanantsoa and Alain Durey.
\newblock Towards a stroke construction model.
\newblock \emph{The International Journal of Table Tennis Sciences}, 1994.

\bibitem[Salimans et~al.(2017)]{SHCSS2017}
Tim Salimans et~al.
\newblock Evolution strategies as a scalable alternative to reinforcement
  learning.
\newblock \emph{arXiv:1703.03864}, 2017.

\bibitem[Schulman et~al.(2017)Schulman, Wolski, Dhariwal, Radford, and
  Klimov]{SWDRK2017arxiv}
John Schulman, Filip Wolski, Prafulla Dhariwal, Alec Radford, and Oleg Klimov.
\newblock Proximal policy optimization algorithms.
\newblock \emph{arXiv:1707.06347}, 2017.

\bibitem[Song et~al.(2020)Song, Gao, Yang, Choromanski, Pacchiano, and
  Tang]{Song2020}
Xingyou Song, Wenbo Gao, Yuxiang Yang, Krzysztof Choromanski, Aldo Pacchiano,
  and Yunhao Tang.
\newblock {ES-MAML:} simple hessian-free meta learning.
\newblock In \emph{ICLR}, 2020.

\bibitem[Sun et~al.(2011)Sun, Xiong, Zhu, Wu, and Chu]{Sun2011BalanceMG}
Yichao Sun, Rong Xiong, Qiuguo Zhu, Jingjing Wu, and Jian Chu.
\newblock Balance motion generation for a humanoid robot playing table tennis.
\newblock \emph{IEEE-RAS Humanoids}, 2011.

\bibitem[Sutton and Barto(1998)]{SB1998MIT}
Richard Sutton and Andrew Barto.
\newblock \emph{Reinforcement Learning: An introduction}.
\newblock MIT Press, MA, 1 edition, 1998.

\bibitem[Sutton et~al.(2000)Sutton, McAllester, Singh, and
  Mansour]{SMSM2000NIPS}
Richard Sutton, David McAllester, Satinder Singh, and Yishay Mansour.
\newblock Policy gradient methods for reinforcement learning with function
  approximation.
\newblock In \emph{NeurIPS}, 2000.

\bibitem[Tebbe et~al.(2018)Tebbe, Gao, Sastre-Rienietz, and Zell]{Tebbe2018ATT}
Jonas Tebbe, Yapeng Gao, Marc Sastre-Rienietz, and Andreas Zell.
\newblock A table tennis robot system using an industrial kuka robot arm.
\newblock \emph{GCPR}, 2018.

\bibitem[Tobin et~al.(2018)]{tobin}
Josh Tobin et~al.
\newblock Domain randomization and generative models for robotic grasping.
\newblock In \emph{IROS}, 2018.

\bibitem[van~den Oord et~al.(2016)van~den Oord, Kalchbrenner, Espeholt,
  kavukcuoglu, Vinyals, and Graves]{PixelCNN}
Aaron van~den Oord, Nal Kalchbrenner, Lasse Espeholt, koray kavukcuoglu, Oriol
  Vinyals, and Alex Graves.
\newblock Conditional image generation with pixelcnn decoders.
\newblock In \emph{NeurIPS}, 2016.

\bibitem[Wierstra et~al.(2008)Wierstra, Schaul, Peters, and
  Schmidhuber]{WSPS2008EC}
Daan Wierstra, Tom Schaul, Jan Peters, and J\"{u}rgen Schmidhuber.
\newblock Natural evolution strategies.
\newblock In \emph{2008 IEEE Congress on Evolutionary Computation}, 2008.

\bibitem[Williams(1992)]{W1992ML}
Ronald Williams.
\newblock Simple statistical gradient-following algorithms for connectionist
  reinforcement learning.
\newblock \emph{Machine Learning}, 1992.

\bibitem[Zhu et~al.(2018)Zhu, Zhao, Jin, Wu, and Xiong]{Zhu2018TowardsHL}
Yifeng Zhu, Yongsheng Zhao, Lisen Jin, Jingjing Wu, and Rong Xiong.
\newblock Towards high level skill learning: Learn to return table tennis ball
  using monte-carlo based policy gradient method.
\newblock \emph{IEEE International Conference on Real-time Computing and
  Robotics}, 2018.

\end{thebibliography}

\end{document}